\documentclass[lettersize,journal]{IEEEtran}
\usepackage{amsmath,amsfonts}
\usepackage{algorithmic}
\usepackage{array}
\usepackage{textcomp}
\usepackage{stfloats}
\usepackage{url}
\usepackage{verbatim}
\usepackage{graphicx}
\graphicspath{ {./images/} }
\usepackage{epstopdf}
\usepackage[table]{xcolor}
\def\BibTeX{{\rm B\kern-.05em{\sc i\kern-.025em b}\kern-.08em
    T\kern-.1667em\lower.7ex\hbox{E}\kern-.125emX}}
\usepackage{balance}
\usepackage[labelformat=simple]{subcaption}

\usepackage{multirow}
\usepackage{booktabs}

\begin{document}
\title{Towards Continual Egocentric Activity Recognition: A Multi-modal Egocentric Activity Dataset\\ for Continual Learning}
\author{
Linfeng~Xu,~Qingbo~Wu,~Lili~Pan,~Fanman~Meng,~Hongliang~Li,~Chiyuan~He,~Hanxin~Wang,~Shaoxu~Cheng,~Yu~Dai
\thanks{L. Xu, Q. Wu, L. Pan, F. Meng, H. Li, C. He, H. Wang, S. Cheng, and Y. Dai are with School of Information and Communication Engineering, University of Electronic Science and Technology of China, Chengdu, China (e-mail: \{lfxu, qbwu, lilipan, fmmeng, hlli\}@uestc.edu.cn).

This work has been submitted to the IEEE for possible publication. Copyright may be transferred without notice, after which this version may no longer be accessible.}% <-this % stops a space
}

\maketitle

\begin{abstract}
With the rapid development of wearable cameras, a massive collection of egocentric video for first-person visual perception becomes available. Using egocentric videos to predict first-person activity faces many challenges, including limited field of view (FoV), occlusions, and unstable motions. Observing that sensor data from wearable devices facilitates human activity recognition (HAR), activity recognition using multi-modal data is attracting increasing attention. However, the deficiency of related dataset hinders the development of multi-modal deep learning for egocentric activity recognition. Nowadays, deep learning in real world has led to a focus on continual learning that often suffers from catastrophic forgetting. But the catastrophic forgetting problem of continual learning for egocentric activity recognition, especially in the context of multiple modalities, remains unexplored due to unavailability of dataset. In order to assist this research, in this paper, we present a multi-modal egocentric activity dataset for continual learning named UESTC-MMEA-CL, which is collected by self-developed glasses integrating a first-person camera and wearable sensors. It contains synchronized data of videos, accelerometers, and gyroscopes, for 32 types of daily activities, performed by 10 participants wearing the glasses. The collection device and process of our dataset are described. Its class types and scale are compared with other publicly available multi-modal datasets for egocentric activity recognition. The statistical analysis of the sensor data is given to show the auxiliary effects for different behaviors. And results of egocentric activity recognition are reported when using separately, and jointly, three modalities: RGB, acceleration, and gyroscope, on a base multi-modal network architecture. To explore the catastrophic forgetting in continual learning tasks on UESTC-MMEA-CL, four baseline methods are extensively evaluated with different multi-modal combinations. We hope the UESTC-MMEA-CL dataset can promote future studies on continual learning for first-person activity recognition in wearable applications. Our dataset will be released soon.
\end{abstract}


\section{Introduction}
\IEEEPARstart{O}{ver} the last decades, enormous annotated images and videos boost the tremendous progress of the models and systems in deep learning and computer vision. Most of popular image and video datasets \cite{Deng09,Everingham15,Lin14,Liu20,Heilbron15} capture moments from a third-person ``spectator" view, which leads to the limited visual perception in current models and systems \cite{Grauman22}. Compared to the widespread third-person images, videos from the egocentric point of view can provide the first-person experience of immersion or ``participant", i.e., we can feel what a person sees when doing an action. Recently, with the rapid development of wearable devices, especially portable head-mounted cameras, such as GoPro, Insta360, Envision Glasses, Vuzix Blade, ThinkReality A3, and Mijia Glasses, the collection of rich egocentric videos becomes available. Analyzing and understanding the content in the egocentric perspective is key to the paradigm shift from ``spectator" view to ``participant" view in computer vision research, which is of prevalent interest due to a large number of applications, including military operations, lifestyle analysis \cite{Cartas20}, human-object interactions \cite{Nagarajan19}, medical monitoring \cite{Zuo18}, augmented and virtual reality \cite{Ng20,Jiang21}, industrial robotics \cite{Smith20}, and autonomous driving \cite{Li22}.

Modeling human activity recognition or anticipation for egocentric videos poses lots of challenges. Firstly, unlike third-person video with apparent motion cues \cite{Su16}, egocentric video changes quickly with the movements of the wearer's head and body. It is difficult to capture the motion cues in egocentric video due to drastic alteration of motion direction and speed, as well as the absence of static backgrounds. Secondly, whereas third person images and videos are captured by a ``spectator" for some purpose, egocentric images are driven by the active behavior of the camera wearer. The attention of the egocentric video or the ``participant", when doing an action, may focus on hands, objects, and the interaction with the surroundings \cite{Li21}, which is quite different from the interest points of a photographer watching from a ``spectator" view. Finally, in some egocentric scenes (e.g., riding bicycle and walking on the road), the objects or body associated with the behavior may not appear in the video due to the limited FoV of the egocentric camera.

Complementary to vision data, inertial sensor data (e.g., gyroscopes and accelerometers) provide position and direction information of the wearable device, which may facilitate human activity recognition for egocentric videos. Recently, with the advancement and application of wearable inertial sensors, multi-modal methods, i.e., combining vision data and sensor data to recognize human activities, are of widespread interest, which may promote vision-based methods \cite{Hu2022, Song16, Rezaie15}. Some pioneering work \cite{Song16} uses LSTM to learn the feature from sensor data and CNNs to learn the feature from vision data, which are fused together to predict wearer's activity. However, due to the difficulty of collecting data and lack of dataset, the progress of multi-modal egocentric activity recognition is slow compared with the vision-based methods.

Nowadays, deep neural networks (DNNs) have made a tremendous progress in various fields and applications, such as computer vision, pattern recognition, and natural language processing. Although this progress is wonderful, most of current DNNs only are good at dealing with static data, because they no longer learn after a training period. This learning strategy is different from what human beings do. Actually, in real world, humans keep acquiring new skills and knowledge to adapt dynamic environments based on what previously learned. This on-going ability is crucial for the development of artificial general intelligence (AGI) \cite{Goertzel07}. However, when the network is trained on a sequence of multiple tasks, the performance on previous tasks will severely degrade, because the weights of the network, which are important for previous tasks, are modified to fit the objectives of the new task. This phenomenon is termed catastrophic forgetting \cite{McCloskey89, ROBINS95}, which has seriously hampered further development of machine learning in real world. To alleviate catastrophic forgetting, many promising continual learning algorithms were proposed in recent years. Most of continual learning research focuses on incremental classification tasks \cite{Li18, Rebuffi17, Hou19, Douillard20, Hu21, Yan21}. To tackle computer vision tasks, continual learning for object detection \cite{Shmelkov17, Joseph21}, semantic segmentation \cite{Michieli19, Douillard21}, and activity recognition \cite{Park21} has attracted much attention and is an emerging trend due to lots of real-world applications, such as robotics and autonomous driving. However, the catastrophic forgetting in the context of continual learning for multi-modal egocentric activity recognition and possible approaches to address this problem have remained unexplored due to unavailability of related dataset. To fill in this gap, we propose a multi-modal egocentric activity dataset for continual learning named UESTC-MMEA-CL.

\begin{figure}
        \centering
        \includegraphics[width=9.7cm]{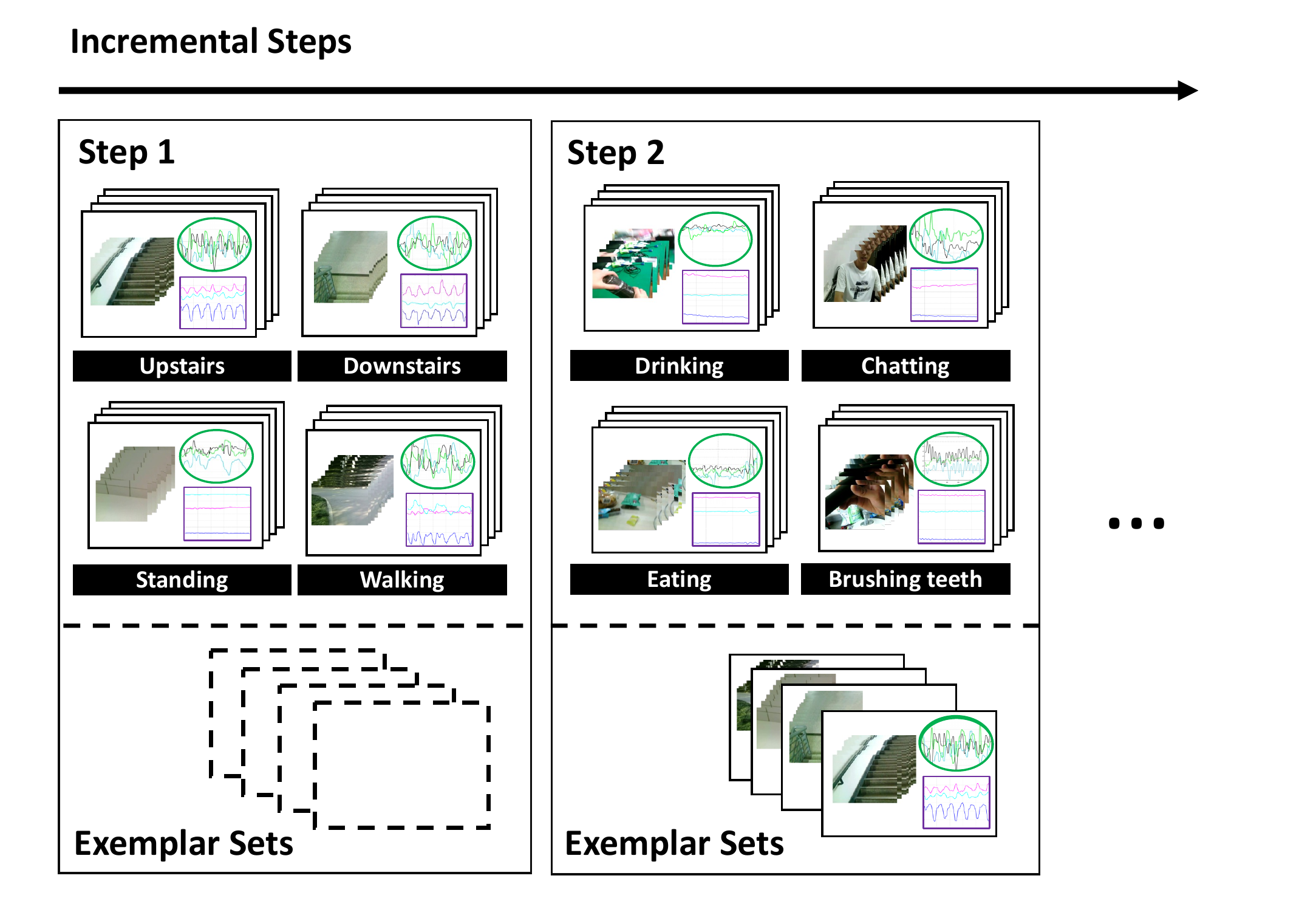}
        \vspace{-1em}
        \caption{Continual egocentric activity recognition with multi modalities: Video stream, acceleration data(green) and gyroscope data(purple).}
        \label{intro_fig} 
\end{figure}

Different from the existing multi-modal egocentric activity datasets \cite{Nakamura17, Spriggs09}, which are collected by separate camera and sensors, our dataset is collected by self-developed glasses integrated with a first-person camera and an inertial measurement unit (IMU). So UESTC-MMEA-CL is suitable to develop applications for life-logging wearable devices (e.g., smart glasses). The vision data and sensor data of our dataset are synchronized well when doing actions. Similar to our manner of collection, MEAD \cite{Song16} was collected by Google Glasses to capture synchronous video and sensor data. However, the scale of MEAD is too limited to take full advantage of DNNs, let alone facilitate research of continual learning. The proposed UESTC-MMEA-CL contains 32 daily activity classes with the duration over 30 hours in total. Each sample clip consists of video, acceleration and gyroscope signals which can provide rich object and motion attributes. Besides, as shown in Fig.~\ref{intro_fig}, we divide these classes into different tasks/steps to adapt to the requirements of continual learning to encourage more research on continual multi-modal egocentric activity recognition.

In order to better describe catastrophic forgetting in the context of continual learning for multi-modal egocentric activity recognition, we propose a benchmark model and evaluate several classic continual learning methods on our UESTC-MMEA-CL.

In summary, the main contributions of this paper are listed as follows:

\begin{itemize}
  \item We propose a new multi-modal egocentric activity dataset UESTC-MMEA-CL, which aims at addressing the catastrophic forgetting problem in the context of continual egocentric activity recognition. To the best of our knowledge, this is the first multi-modal dataset for continual egocentric activity recognition.
  \item We propose a benchmark model for mulimodal egocentric activity recognition and demonstrate the experimental results when using separately, and jointly, the three modalities, i.e., RGB, acceleration, and gyroscope, on UESTC-MMEA-CL. 
  \item We set the continual egocentric activity recognition tasks and describe the main challenges raised by UESTC-MMEA-CL: the catastrophic forgetting of each modality. Besides, we try to employ popular continual learning methods to tackle this problem and provide some potential research directions.
\end{itemize}

\section{Related Work}
\subsection{Multi-modal Human Activity Recognition}
\subsubsection{Datasets}
In order to integrate the complementary information for vision data, some multi-modal datasets have been proposed for human activity recognition task. UTD-MHAD \cite{Chen15} is collected from two independent devices, i.e., a Kinect camera and a wearable inertial sensor. The dataset consists of RGB videos, depth videos, skeleton positions, and inertial signals for 27 human actions such as right arm throw, cross arms in the chest, basketball shoot, et al. For the purpose of developing and evaluating multi-modal algorithms, Berkeley-MHAD \cite{Ofli13} consists of multi-modal data for 11 actions, which is captured by five different systems: an optical motion capture system, stereo cameras, Microsoft Kinect cameras, accelerometers, and microphones. To address the health problem of elder persons, the Up-Fall dataset \cite{Martinez19} is proposed for reliable fall detection. The dataset contains multi-modal data for six daily living activities and five types of simulated falls from wearable sensors, ambient sensors, and vision devices.

Nowadays, activity recognition from the egocentric perspective has become a widely concerned topic due to the interesting life-logging applications, such as lifestyle analysis and health monitoring \cite{Song16}. However, the progress of multi-modal egocentric activity recognition is relatively slow because it is not easy to capture multi-modal data from wearable devices such as smart glasses. The existing multi-modal datasets for egocentric activity recognition are quite limited. EPIC-KITCHENS \cite{Damen22} is a large-scale egocentric video dataset which is collected by 32 participants in kitchen environments. Every participant is commanded to use a head-mounted GoPro Hero7 black to record every second from the time they entered the kitchen. This dataset contains multi-modal data of RGB, flow, and audio, except position and direction information related to the activities. With the wide use of wearable sensors, a number of works introduced some auxiliary data for a comprehensive understanding of human activities. Stanford-ECM \cite{Nakamura17} consists of egocentric video, accelerometer data, and heart rate data, collected by a mobile phone placed in the chest pocket and a wrist-worn heart rate sensor. The dataset contains 24 daily activities under natural conditions, including various levels of motion intensity. CMU-MMAC \cite{Spriggs09} introduces 29 kitchen activities, such as opening fridge, removing cap, which is collected from 7 participants using an egocentric camera, IMUs, and other sensors. In the existing datasets, the MEAD dataset collected by Song et al. \cite{Song16} is most similar to our proposed UESTC-MMEA-CL dataset. The MEAD dataset was collected by Google Glasses to record 20 human activities which contains modalities of synchronous video and sensor data. However, there are only 200 sequences in total in the MEAD dataset, whose scale is too limited for the research of DNNs.

\subsubsection{Methods}
Datasets with more complex scenes and more categories of behaviors make it challenging to recognize human activities for vision-based methods. Is is helpful to integrate complementary information for vision data. In order to improve the algorithm's robustness, Song et al. \cite{Song16} disassemble the visual signals into three input forms (single frame, optical flow, and stabilized optical flow), then classify activities with the aid of gyroscope and accelerometer data. Kazakos et al. \cite{Kazakos19} propose a mid-level fusion Temporal Binding Network (TBN) to combine signals of three modalities, i.e., video, flow, and audio. Different from traditional fusion method, multi-modal signals are aggregated before temporal fusion with the shared weights over time and each modality is trained individually. Spriggs et al. \cite{Spriggs09} segment human motion into several actions and classify activities for first-person sensing, which is captured by a wearable vision sensor and IMUs. Kitani et al. \cite{Kitani11} propose an unsupervised method for the egocentric activity recognition task, which adopts a stacked Dirichlet process mixture model to infer the motion histogram codebook and the activity category. Nakamura et al. \cite{Nakamura17} employ a stacked LSTM network to process the fused features from vision and acceleration, then jointly predict activities and energy expenditures with the aid of heart-rate sensor data. Besides, some researches \cite{Li13, Bambach15} have been devoted to predicting people’s intentions by analyzing some mid-level features like people’s face, gaze, and hands.

\subsection{Continual Learning}
\subsubsection{Datasets}
To the best of our knowledge, there are no datasets dedicated to continual learning tasks, and the researchers usually manually divide the well-known datasets into continual learning task sequences according to the specific task types, such as image classification, object detection and image segmentation, etc. Specifically, for the image classification task, the most widely-used datasets are ImageNet \cite{Deng09} and CIFAR100 \cite{Krizhevsky09}, which are originally used for non-continuous image classification. ImageNet consists of 1000 classes with approximately 1000 pictures for each class. The size of each picture is $224 \times 224$. CIFAR100 is made up of 60000 images evenly divided into 100 classes, where each class is comprised of 500 training samples and 100 test samples. For the image segmentation task, the selected datasets are Pascal-VOC 2012 \cite{pascal-voc-2012} and ADE20K \cite{Zhou17}. The former contains 20 classes and the latter contains 150 classes. The Pascal-VOC 2012 dataset is also widely used in the object detection task and the action classification task. Another popular dataset in object detection is Microsoft-COCO \cite{Lin14}, which contains 80 classes in total and is comprised of more than 300,000 images and more than 2 million instances. It is worth noting again that all the datasets mentioned here are used originally for non-continuous tasks, but the researchers in continual learning manually partition them into continuous task sequences.

\subsubsection{Methods}
Many efforts have been made to improve the performance of continual learning. The existing work can be mainly divided into parameter-based, knowledge-distillation-based, and parameter-expansion-based.

\textbf{Parameter-based.} The key to this method is to evaluate the importance of parameters and protect the important ones. Methods \cite{Kirkpatrick17, Zenke17, Aljundi18, Liu18} fall into this category with different parameter importance estimations. A quadratic penalty imposed on the parameters critical to old tasks is proposed by EWC \cite{Kirkpatrick17}. The authors utilize the Fisher Information Matrix \cite{MYUNG03} to choose the critical parameters. Liu et al. \cite{Liu18} obtain a better Fisher Information Matrix approximation by rotating the parameter space.

However, overestimation and underestimation might happen due to batch updates. To solve this problem, \cite{Zenke17} accumulates the changes in the learning of parameters via which the importance is estimated. Memory Aware Synapses (MAS) \cite{Aljundi18} solves the same problem by accumulating the gradient magnitude.

\textbf{Distillation-based.} The core idea of this category is to prevent the drift between new and old models. Learning without Forgetting (LwF) \cite{Li18} first introduces knowledge distillation to continual learning. Specifically, the predictions made by the new model should be close enough to the old model predictions. iCaRL \cite{Rebuffi17} proposes a rehearsal strategy and a nearest-mean-of-exemplars classifier to cooperate with the LwF loss. The less-forget loss is then devised by UCIR \cite{Hou19}, which penalizes the activation drift of the backbone. For a stronger distillation constraint, a spatial-based multi-level distillation loss is designed by PODNet \cite{Douillard20}. DDE \cite{Hu21} aims to solve catastrophic forgetting from the scope of causal analysis and then proposes to distill the colliding effect of new and old data.

\textbf{Parameter-expansion-based.} Another straightforward idea is to prevent the parameters of previous tasks from drifting and expand new branches for new tasks. EG \cite{Aljundi17} allocates a duplicate model for new tasks. CCGN \cite{Abati20} devises a task-specific gating mechanism to select the target filters for specific inputs. DER \cite{Yan21} also duplicates the entire backbone to learn new classes. Additionally, DER concatenates all the features obtained from the backbones and utilizes them to learn a unified classifier. However, the excessive parameter overhead hinders the application of these methods in real-world scenarios.

\section{UESTC-MMEA-CL Dataset}

In this section, we introduce the data collection of UESTC-MMEA-CL, present statistics, and compare with other multi-modal egocentric datasets. The distributions of standard deviation of acceleration and gyroscope sensor data are shown to demonstrate the motion intensity of each activity, as well as the motion correlation of the two sensor modalities. 

\subsection{Data Collection}

\begin{figure}
    \centering
    \begin{subfigure}{.22\textwidth}
        \centering
        \includegraphics[width=\textwidth]{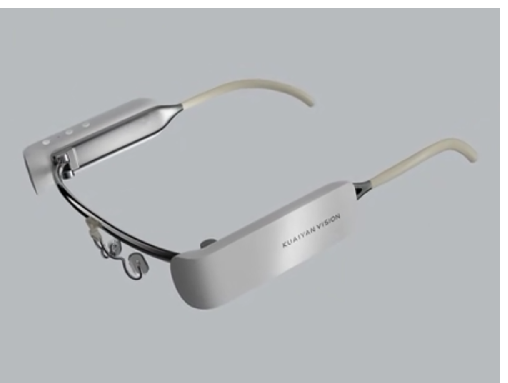}
        \caption{}
        \label{fig:1:a} %% label for second subfigure
    \end{subfigure}
    \begin{subfigure}{.22\textwidth}
        \centering
        \includegraphics[width=\textwidth]{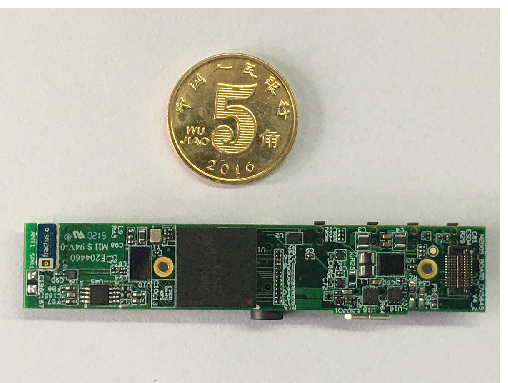}
        \caption{}
        \label{fig:1:b} %% label for second subfigure
    \end{subfigure}
    \caption{The device for data collection. (a) Our developed Kuaiyan Vision Smart Glasses. (b) The mainboard of the glasses.}
 \end{figure}

In order to collect synchronous video and sensor data for egocentric activity recognition, we developed a pair of wearable smart glasses, as show in Fig.~\ref{fig:1:a}, with a first-person camera, IMU sensors, and the function of wireless connection. The mainboard of the glasses is very tiny as shown in Fig.~\ref{fig:1:b}. The process of data collection can be conducted in the following two steps: 1) device configuration; 2) data collection and post-processing.

First, we set up the device as follows. For camera, the video resolution is $640 \times 480$, and the frame rate is 25FPS. For sensors, the sample rate is 25Hz. The sensitivity of gyroscope is 16.4LSB/deg/s, and the sensitivity of accelerator is 8192LSB/g. We developed applications to capture videos, accelerometers, and gyroscopes data, which are synchronized by time-delay correction and transferred to a terminal via WIFI.

After the configuration, ten subjects are divided into five groups. For each group, one subject equips the glasses and acts, while another uses a terminal to ensure each video only contains one action. All data are collected from different scenes with adequate illumination. Because the sensors are sensitive to noise, the median filtering method is used to filter the abnormal values and noise. The kernel size of the median filter is 5. After filtering the sensor data can reflect the movement of the subject better.

%figure 1

\subsection{Dataset Overview}
\label{Overview}

We first introduce some general statistics of our proposed dataset UESTC-MMEA-CL, compared with the available egocentric datasets, which is shown in Table~\ref{T1}. Our dataset comprises 30.4 hours of video clips, acceleration stream and gyroscope data in total. There are 32 daily activities included in our dataset as shown in Table~\ref{T2}, containing some basic movements (upstairs, walking, standing, etc.), indoor behaviors (writing, reading, type-PC, etc.), some kinds of cleaning labor (mop-floor, wash-dish, wipe-table, etc.), several recreations and leisure activities (watch-TV, play-phone, play-card, etc.), activities with hands (wash-hands, wash-dish, and cooking), and activities with head movements (eating and drinking).

\begin{table*}[h]

\centering
\caption{Comparison with available egocentric datasets.}
\vspace{-0.5em}
\resizebox{\textwidth}{21.5mm}{%
\setlength\tabcolsep{3.6mm}
\begin{tabular}{lccccc|ccc}
\toprule[1pt]
Dataset         & \#Subjects & \#Class & \#Duration (h) & Mount & Scenario & Video & Acc & Gyro \\ \hline
CMU-MMAC \cite{Spriggs09}        & 39       & 29    & 17.0        & Head  & Natural  & \(\checkmark\)     & \(\checkmark\)  & \(\checkmark\)    \\
JPL-Interaction \cite{ryoo2013} & 1        & 7     & 0.4         & Head  & Indoor   & \(\checkmark\)     &     &      \\
MEAD \cite{Song16}   & 7       & 20   &   0.5        & Head  & Natural  &
\(\checkmark\)     &  \(\checkmark\)   &  \(\checkmark\)    \\ 
GTEA Gaze \cite{fathi2012} & 14        & 40     & 1.0         & Head  & Kitchen   & \(\checkmark\)     &     &      \\
GTEA Gaze+ \cite{fathi2012} & 5     & 44     & 9.0         & Head  & Kitchen   & \(\checkmark\)     &     &      \\
PAMAP2  \cite{reiss2012}        & 9        &   18  &      -     &    -  &     -    &       & \(\checkmark\)  &      \\
UEC EgoAction \cite{Kitani11}  & 1        & 37    & 0.5         & Head  & Kitchen  & \(\checkmark\)     &     &      \\
Stanford-ECM  \cite{Nakamura17} & 10       & 24    & 31.0        & Chest & Natural  & \(\checkmark\)     & \(\checkmark\)   &      \\

EPIC-KITCHENS  \cite{Damen22} & 32       & 149   &    -       & Head  & Kitchen  &
\(\checkmark\)     &     &      \\ \hline
UESTC-MMEA-CL(ours)   & 10       & 32    &     30.4        & Head  & Natural  & \(\checkmark\)    & \(\checkmark\)   & \(\checkmark\)    \\ \bottomrule[1pt]
\end{tabular}
}
\footnotesize
\label{T1}
\end{table*}

\begin{table}[]
\caption{Activities in UESTC-MMEA-CL Dataset.}
\vspace{-0.5em}
\centering
\label{T2}
\resizebox{0.5\textwidth}{50mm}{
\setlength{\tabcolsep}{0.46mm}
\begin{tabular}{l|lccl}
\toprule[1pt]
 & Class &\#Clips  & \#Avg-Dur(s) & Scenario \\ \hline
 1& upstairs &192  &17.7  & teaching building, park, library \\
 2& downstairs &190  &17.3  & teaching building, park, library \\
 3& drinking &202  &16.0  & dorm, office \\
 4& fall &185  &13.7  &  campus, office, corridor\\
 5& reading &201  &18.2  &  office, classroom\\
 6& sweep-floor &229  &18.0  &  corridor, office, campus\\
 7& cut-fruits &203  &17.5  &  teaching building, park, office\\
 8& mop-floor &206  &17.7  & corridor, office\\
 9& writing &209  &18.8  &  classroom, office\\
 10&wipe-table  &245  &18.2  & home, office, dorm \\
 11&wash-hand  & 189 &17.0  &  bathroom, kitchen\\
 12&standing  &203  &18.0 & corridor, office, dining hall\\
 13&play-phone  &205  &17.4  & classroom, office, campus, park \\
 14&type-PC  &204  &18.1  & classroom, office \\ 
 15&eating  & 213 &17.1  & classroom, office, dining hall, canteen \\ 
 16&cooking  & 225 & 17.1 & kitchen, office \\
 17&pick-up-phone  &213  &14.4  & classroom, office, teaching building, campus \\
 18&drop-trash  &201  &13.1  & campus, park, teaching building  \\
 19&fold-clothes  &204  &17.3  &home, dorm, office  \\
 20&walking  &203  &17.1  & campus, library, park \\
 21&play-card  &206  & 17.0 &  classroom, restroom\\
 22&brush-teeth  &203  &17.0  &  bathroom\\
 23&wash-dish  &189  & 16.0 &  kitchen, bathroom\\
 24&moving-sth  &201  & 15.7 &  corridor, office, teaching building\\
 25&type-phone  &195 & 17.3 &  classroom, office, campus\\
 26&chat  &203  & 17.3 &  classroom, office, dorm\\
 27&open-close-door  &200  & 15.8 & office, home \\
 28&ride-bike  &198  &17.1  &  campus, park, road\\
 29&sit-stand  &201  &15.7  &  office, classroom, library\\
 30&take-drop-sth  &201  & 13.5 & office, classroom, library\\
 31&shopping  &208  &17.1  &  mall, street\\
 32&watch-TV  &205  & 16.9 &  office, home\\ \bottomrule[1pt]
\end{tabular}%
}

\end{table}

In contrast to Stanford-ECM \cite{Nakamura17}, which suffers from limited FoV and contextual information due to the lower location of chest-mounted camera, we embed the camera into the head-mounted glasses to capture more useful visual information. Compared with uni-modality datasets such as JPL-Interaction \cite{ryoo2013}, GTEA Gaze \cite{fathi2012}, GTEA Gaze+ \cite{fathi2012}, UEC EgoAction \cite{Kitani11}, EPIC-KITCHENS \cite{Damen22}, we provide additional synchronized data of accelerometers and gyroscopes, which are complementary to vision data and make it available to explore the catastrophic forgetting problem using separately, and jointly, the three modalities. CMU-MMAC \cite{Spriggs09} provides multi-modal measures of human activities with four wireless IMUs and five wired IMUs located on multiple parts of the subjects' body, such as wrists, ankles, arms, and waist, in order to capture motion details when performing cooking and food preparation. However, the complex devices make the data collection of daily behaviors difficult, which is not suitable for wearable applications. MEAD \cite{Song16} contains 20 life-logging activities, which uses Google Glasses to capture multi-modal data of video, accelerometer and gyroscope. But due to the limited scale, duration, and category number of MEAD, it is difficult to take full advantage of DNNs and set up multiple tasks for continual learning research. 

\subsection{Dataset Statistics}
\label{Dataset Statistics}
\begin{figure}
        \centering
        \includegraphics[width=8.6cm]{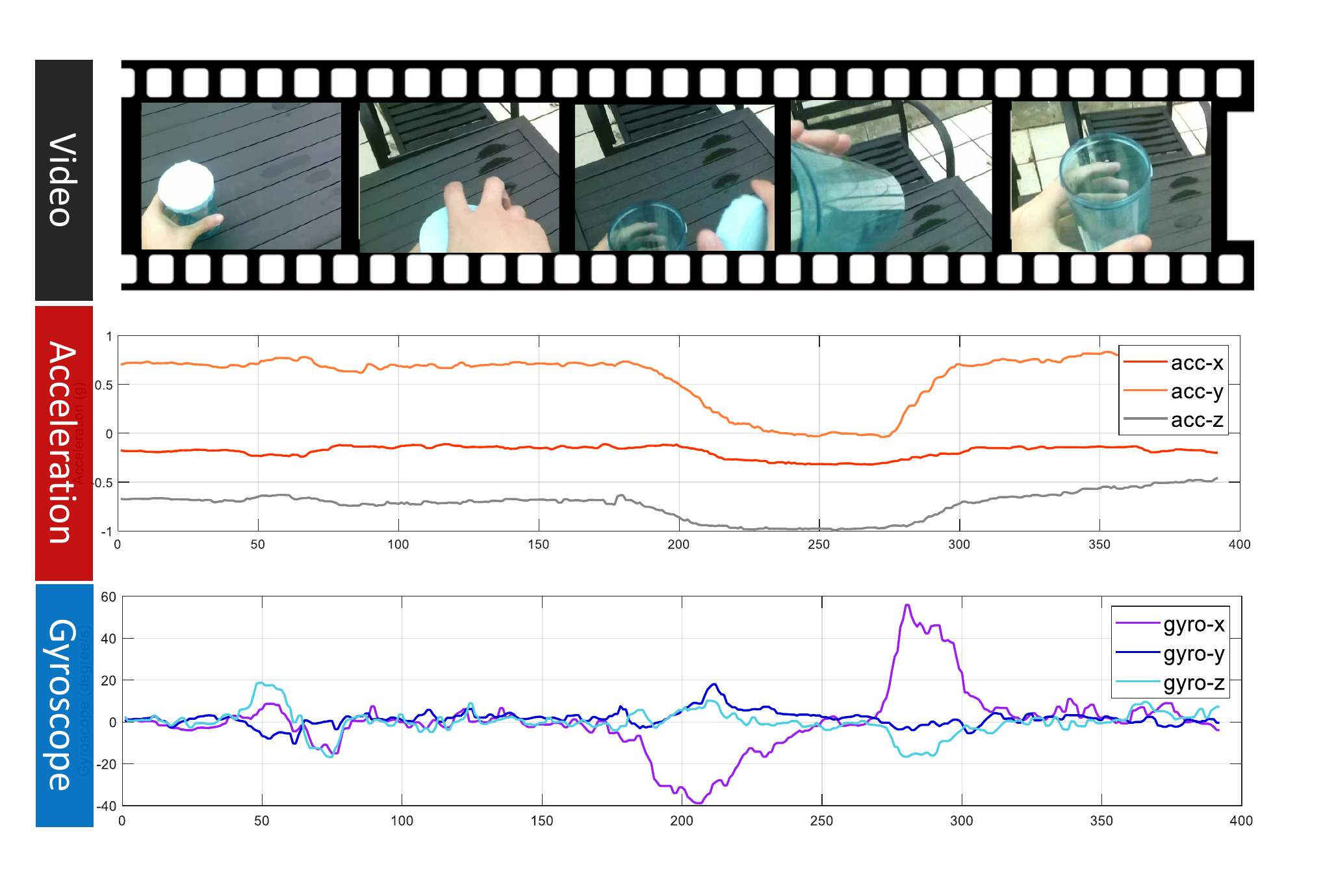}
        \vspace{-1em}
        \caption{A sample of activities “drinking”, which consists of the synchronized video stream, acceleration, and gyroscope sensor data.}
        \label{sample} 
\end{figure}

Our UESTC-MMEA-CL contains 32 different activity classes and each class contains approximately 200 samples, consisting of fully synchronized first-person video clips, acceleration sensing sequences, and gyroscope sensing sequences. A sample is shown in Fig.~\ref{sample}.

\begin{figure*}
        \centering
        \includegraphics[width=18cm]{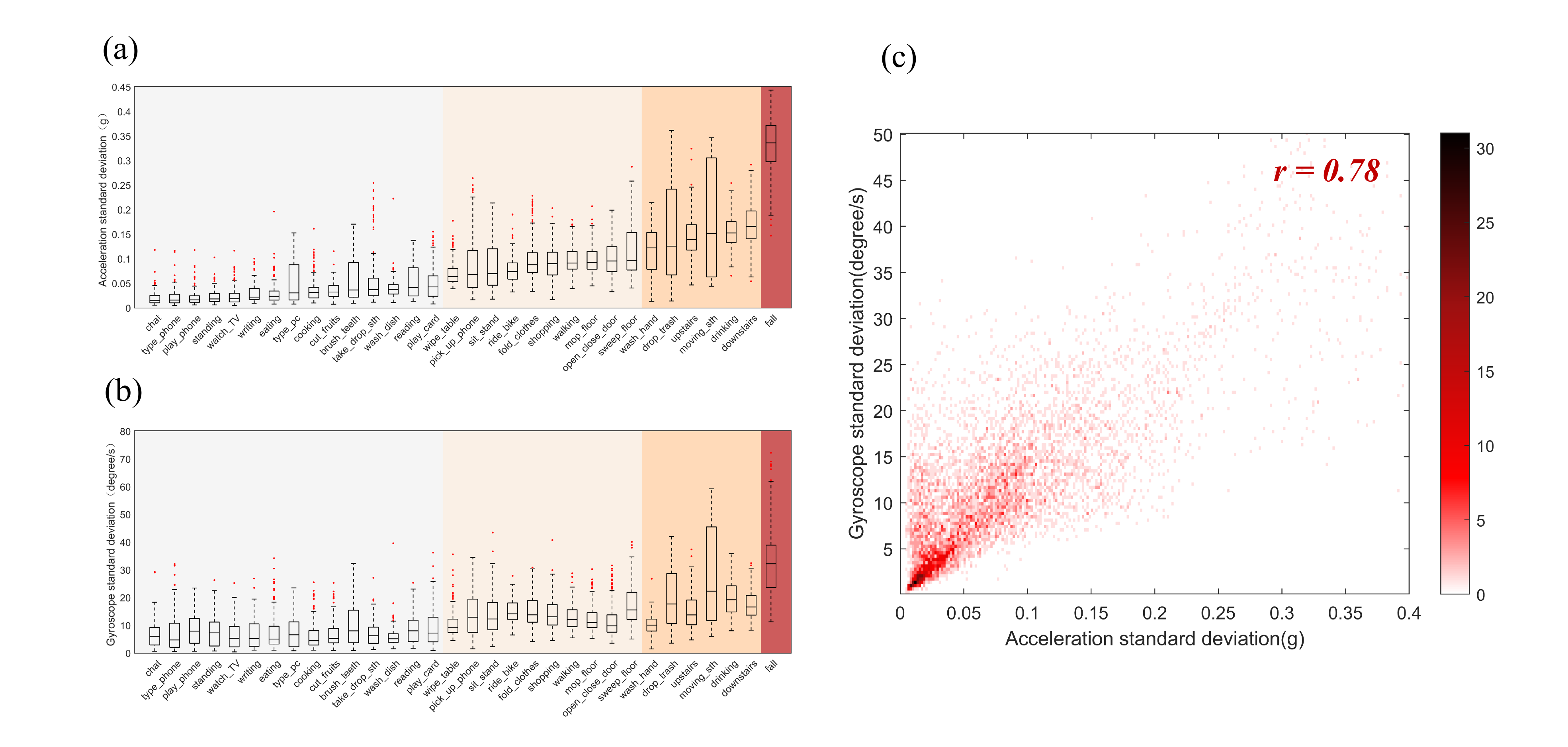}
        \vspace{-1em}
        \caption{Statistics of sensor data. (a) STD distributions of acceleration for all activity classes. The relative motion intensity of the activities increase sequentially from the leftmost column to the right, which are divided into four different levels according to the median STD. (b) STD distributions of gyroscope for each activity. (c) Scatter plot of the STD distributions of acceleration and gyroscope (Correlation coefficient \(r = 0.78\) on all samples).
}
        \label{statis} 
\end{figure*}

Although visual information dominates human activity recognition, sensor data may provide complementary position and direction information to facilitate the recognition task for egocentric video. In order to demonstrate the auxiliary motion information of the sensor data, we make statistical analysis of the two modalities, i.e., accelerometer and gyroscope data. Following Stanford-ECM \cite{Nakamura17}, we calculate the standard deviation (STD) of the sensor data to show the relative motion intensity for each activity. Fig.~\ref{statis}(a) demonstrates the distribution of acceleration STD. Activities are sorted by the median STD of acceleration and divided into four levels of intensity. Behaviors such as upstairs, downstairs, and moving-sth are relatively vigorous while chat, type-phone, and watch-TV are stable. The degree of variation in orientation can be measured by the STD of the gyroscope data, which is shown in Fig.~\ref{statis}(b). Besides, Fig.~\ref{statis}(c) shows a scatter plot of the STD distributions of acceleration and gyroscope data, which reflects the motion correlation (correlation coefficient \(r = 0.78\)) of the acceleration and gyroscope data.

\section{Continual learning on UESTC-MMEA-CL}
\label{Continual learning on UESTC-MMEA-CL}

 \subsection{Problem Setup}
 \label{problem setup}
Although DNNs have made a remarkable progress in many applications, most of current DNNs face the catastrophic forgetting problem when dealing with dynamic data. In the wearable application of egocentric activity recognition, data may come dynamically. Limited by the memory capacity and computing power of wearable devices, models are expected to accommodate new recognition tasks when the data from the past are inaccessible or partially accessible. In order to explore the catastrophic forgetting and promote possible approaches to address this problem, we introduce continual learning into our task scenario.

The activity class set $\mathcal{C}$, which contains $N$ classes ($N=32$), of our dataset is divided into $S$ incremental steps/tasks. The class set $\mathcal{C}^{s}$ of each step/task $s$ $(0 \le s \le S-1)$ contains $N/S$ classes ($\mathcal{C}^{s}=\bigcup_{l=0}^{(N/S) - 1} \mathcal{C}^{s}_{l} = \{\mathcal{C}^{s}_{0}, \mathcal{C}^{s}_{1}, ..,\mathcal{C}^{s}_{(N/S) - 1}\}$). The multi-modal sample set of class $\mathcal{C}^{s}_{l}$ is denoted as $\mathcal{D}^{s}_{l}=\left\{\left.\left(\mathbf{v}_{i}^{l,s},\mathbf{a}_{i}^{l,s},\mathbf{g}_{i}^{l,s}, \mathbf{y}_{i}^{l,s}\right)\right|_{i=1} ^{N_{l,s}}\right\}$. Sample set $\mathcal{D}^{s}_{l}$ contains $N_{l,s}$ pairs of samples $\mathbf{x}^{l,s}_{i} = (\mathbf{v}_{i}^{l,s}, \mathbf{a}_{i}^{l,s}, \mathbf{g}_{i}^{l,s})$ and activity class label $\mathbf{y}^{l,s}_{i}$, where $\mathbf{v}_{i}^{l,s}$, $\mathbf{a}_{i}^{l,s}$, $\mathbf{g}_{i}^{l,s}$ represent the visual signal, acceleration signal and gyroscope signal of the $i$-th sample of $\mathcal{C}^{s}_{l}$, respectively. At the $s$ step, the models are trained with the available samples $\mathcal{D}^{s}$, where $\mathcal{D}^{s} = \bigcup_{l=0}^{(N/S)-1} \mathcal{D}^{s}_{l}$, and evaluated on the test set of all seen classes $\bigcup_{j=0}^{s} \mathcal{C}^{j}$. For the methods based on exemplar replay (or rehearsal), we define a replay buffer to store exemplar samples $\mathcal{E}^{s}$ of old classes. At the first step, available data is only $\mathcal{D}^{0}$, and at each subsequent incremental step, the available data is $\mathcal{D}^{s} \cup \mathcal{E}^{s} (s\ge 1)$.

\subsection{Base Multi-modal Architecture}
\begin{figure}
        \centering
        \includegraphics[width=9cm]{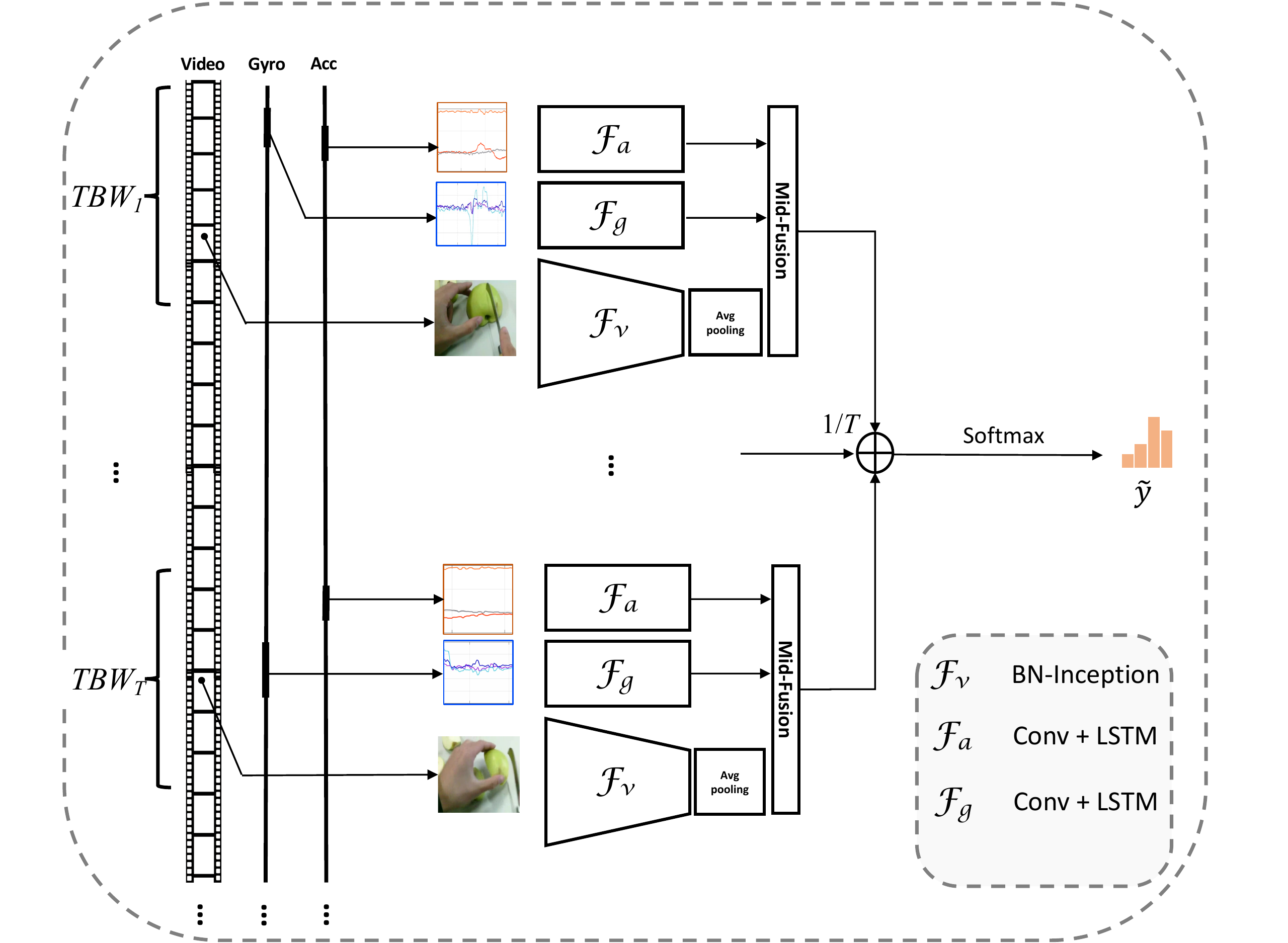}
        \vspace{-1em}
        \caption{Base architecture of multi-modal egocentric activity recognition. The number of TBWs $T$ is set to 8.}
        \label{TBN} 
\end{figure}

First, we introduce the base architecture employed for the multi-modal egocentric activity recognition, which is shown in Fig.~\ref{TBN}. The architecture is based on the temporal binding network (TBN) \cite{Kazakos19}, which is effective for modal fusion and temporal aggregation. BN-Inception \cite{ioffe2015batch} is adopted as the feature extractor $\mathcal{F}_v$ for the frame from the video stream. The deep convolutional and LSTM recurrent neural networks \cite{ordonez2016deep} are used for the feature extraction of acceleration signal $\mathcal{F}_a$ and gyroscope signal $\mathcal{F}_g$. We use the random sampling method to sample the multi-modal data within a temporal blinding window (TBW) \cite{Kazakos19}. The input multi-modal data ${x}=\left \{v, a, g \right\}$, where $v$, $a$, and $g$ denote the video, acceleration data, and gyroscope data respectively, are divided into $T$ TBWs. Within a temporal blinding window $TBW_t$ ($1 \le t \le T$), modalities are sampled as a single video frame, a sequence of acceleration data, and gyroscope data, which is denoted as ${x}_{t}=\left \{ v_t,a_t,g_t \right \}$. Thus, we can get fused feature:

\begin{equation}
y_{t} = Q\left [ p \left ( \mathcal{F}_v \left ( v_{t} \right ) \right ),   \mathcal{F}_a\left ( a_{t} \right ) ,   \mathcal{F}_g\left ( g_{t} \right ) \right ],
\end{equation}

\noindent where $p$ denotes the average pooling operation, and $Q$ represents the mid-fusion block to aggregate features of the three modalities, which contains concatenation, convolution, and ReLU operations. Then, all features $y_{t}$ from the $T$ TBWs are averaged as the input to the activity classifier:

\begin{equation}
\tilde{y} = softmax(\frac{1}{T} \sum_{t=1}^{T} y_{t}).
\end{equation}

\noindent Following most classification tasks, cross entropy is employed as the loss function for the final prediction of activities, and branches of all modalities are trained jointly.
                                                                      
 \subsection{Benchmark for Continual Learning}
 In this paper, we implement three baseline methods, as well as the most straightforward fine-tune solution, on the base multi-modal architecture (Fig.~\ref{TBN}) as the benchmark for continual learning on UESTC-MMEA-CL dataset. These continual learning methods are as follows:

\begin{itemize}
\item \textbf{EWC \cite{Kirkpatrick17}:} a parameter-based continual learning model, where the important parameters to old tasks are regularized and changed in a small range. Therefore the influence to old tasks is alleviated during new task learning.
\end{itemize}

\begin{itemize}
\item \textbf{LwF \cite{Li18}:} a distillation-based continual learning model, where knowledge distillation (KD) is combined with fine-tuning, and the output of the old network is used to constrain the parameter update of the new task. 
\end{itemize}

\begin{itemize}
\item \textbf{iCaRL \cite{Rebuffi17}:} a replay-based continual learning model, which constructs and manages an exemplar set consisting of collection of representative old data. The exemplars that are closest to the mean feature of each class are selected. For the new task, the new data and exemplar set are mixed as input in the learning phase.
\end{itemize}

\section{Experiments}
 \subsection{Implementation Details}
 
\textbf{Sensor signal processing:} We use a median filter with kernel size 5 to filter abnormal values of acceleration and gyroscope signals. Since the gyroscope is not reliable in a long term, the trapezoidal integral of the filtered angular velocity is calculated to get the angle data. However, there exists a bias drift problem in the gyroscope signal which would cause a large cumulative error in the integral results. To tackle this problem, we subtract the mean value before the integral. After the filtering and integral processing, 24 consecutive sensor data are sampled within a TBW.
 
\textbf{Multi-modal training details:} We implement the model in PyTorch. The video stream branch is trained by the SGD optimizer \cite{qian1999momentum} with a momentum of 0.9, a batch size of 8, a dropout of 0.5, and a learning rate of 0.001. The acceleration and gyroscope steam branches are trained by the RMSprop optimizer \cite{hinton2012neural} with a dropout of 0.5 and a learning rate of 0.001. The batch size is set to 32 for the acceleration network and 8 for the gyroscope network. We initialize the RGB network with pre-trained model from the ImageNet. All networks are trained for 50 epochs, and the learning rate is decayed by a factor of 10 at epoch 10 and 20.

\begin{figure*}
        \centering
        \includegraphics[width=16cm]{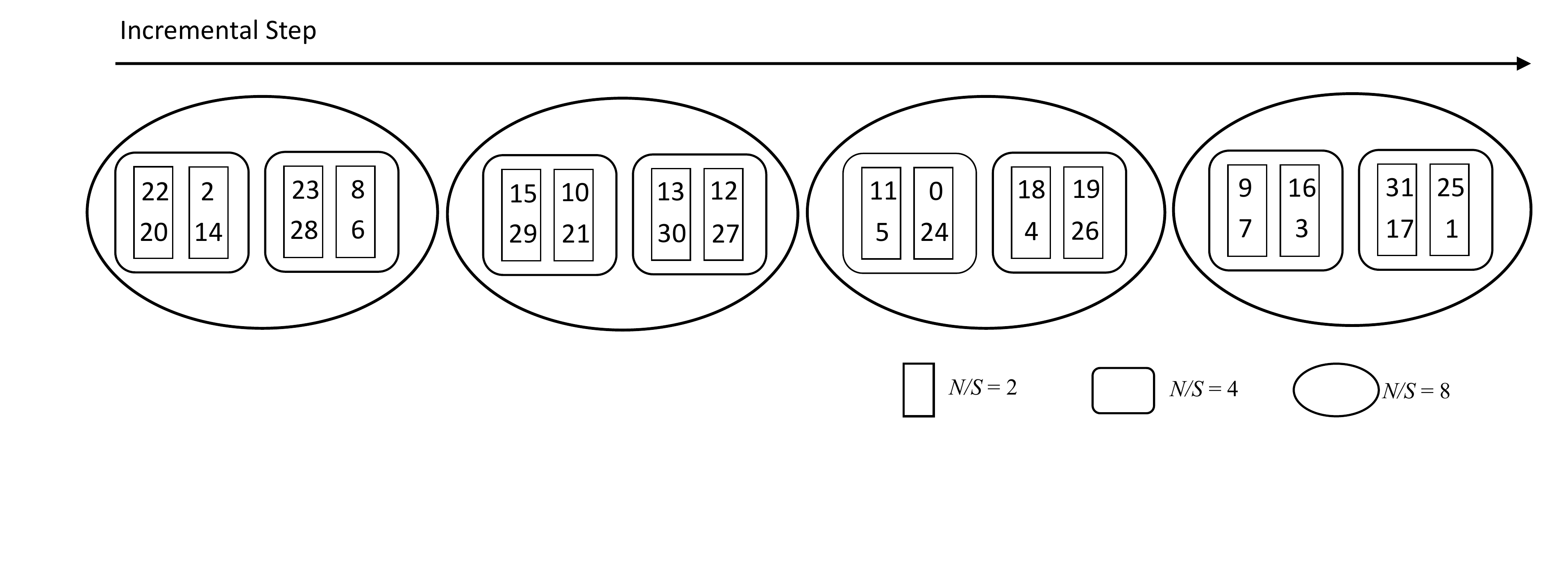}
        \vspace{-4em}
        \caption{Settings of incremental steps. Each number denotes the activity class in Table~\ref{T2}.}
        \label{CL_set} 
\end{figure*}

\textbf{Continual learning training details:} All continual learning benchmarks are implemented using PyTorch and PyCIL \cite{zhou2021pycil}. The settings of incremental steps and activity classes in each step are shown in Fig.~\ref{CL_set}. Based on the Problem Setup introduced in \ref{problem setup}, we set the number of total activity classes $N = 32$ and the number of incremental steps $S = \left \{16, 8, 4 \right\}$. Therefore, each step contains $N/S = \left\{2,4,8\right\}$ activity classes. For the replay-based continual learning method, we set the memory size to 320. Other parameter settings are the same as multi-modal training.

\subsection{Metrics}
\label{metrics}
Following the research in continual learning \cite{Rebuffi17, chaudhry2018riemannian}, two metrics, i.e., average accuracy and average forgetting, are used to evaluate the overall accuracy in continual learning stages and the average decrease of accuracy on previous tasks, respectively, which is defined as follows.

\textbf{Average accuracy (A)} Here, $a_{k,j} \in [0,1]$ denotes the accuracy evaluated on the test set of task $j$ after learning task $k$ ($j \leq k$). Then the average accuracy on task $k$ can be calculated as
 
\begin{equation}
A_k = \frac{1}{k} \sum^k_{j=1} a_{k,j}
\end{equation}
 
\textbf{Average forgetting (F)} The forgetting for a certain task is defined as the difference between the maximum knowledge obtained with respect to the task during the learning process in the past and the current knowledge the model has about it \cite{chaudhry2018riemannian}. $f^k_j \in [-1,1]$ denotes the forgetting on the previous task $j$ after learning task $k$, which can be formulated as 
\begin{equation}
f^k_j = \max_{l \in {j, \cdots, k-1}} a_{l,j}-a_{k,j},  \quad\forall j<k
\end{equation}

\noindent Thus, the average forgetting at the $k$-th task can be defined as 
\begin{equation}
F_k = \frac{1}{k-1} \sum^{k-1}_{j=1} f^k_j
\end{equation}

\noindent Note that the lower $F_k$, the less forgetting of a model on the previous tasks.

 \subsection{Evaluation on UESTC-MMEA-CL}

\begin{table*}[]
\caption{Results on UESTC-MMEA-CL using multi-modal combinations (`All' denotes `RGB+Acc+Gyro').}
\vspace{-0.5em}
\centering
\label{T3}
\resizebox{\textwidth}{9.1mm}{
\setlength{\tabcolsep}{5mm}
\begin{tabular}{l|lll|cccc}
\toprule[1pt]
\multirow{2}{*}{}   & \multicolumn{3}{c|}{Uni-modal} & \multicolumn{4}{c}{Multi-modal} \\ \cline{2-8} 
 & \multicolumn{1}{l}{RGB} & \multicolumn{1}{l}{Acc} & Gyro & \multicolumn{1}{l}{RGB + Acc} & \multicolumn{1}{l}{RGB + Gyro} & \multicolumn{1}{l}{Acc + Gyro} & All \\ \hline
Top1-Accuracy (\%)       & 92.6    &   35.0     &    38.2    &  94.5    & 93.9     & 59.7     & \textbf{95.6}    \\
Avg Class Precision (\%) &  92.5      &  35.1      &  38.3      &  94.4    &   93.9   & 59.9     &  \textbf{95.6}    \\ 
\bottomrule[1pt]
\end{tabular}
}
\end{table*}

\begin{figure*}
        \centering
        \includegraphics[width=19.5cm]{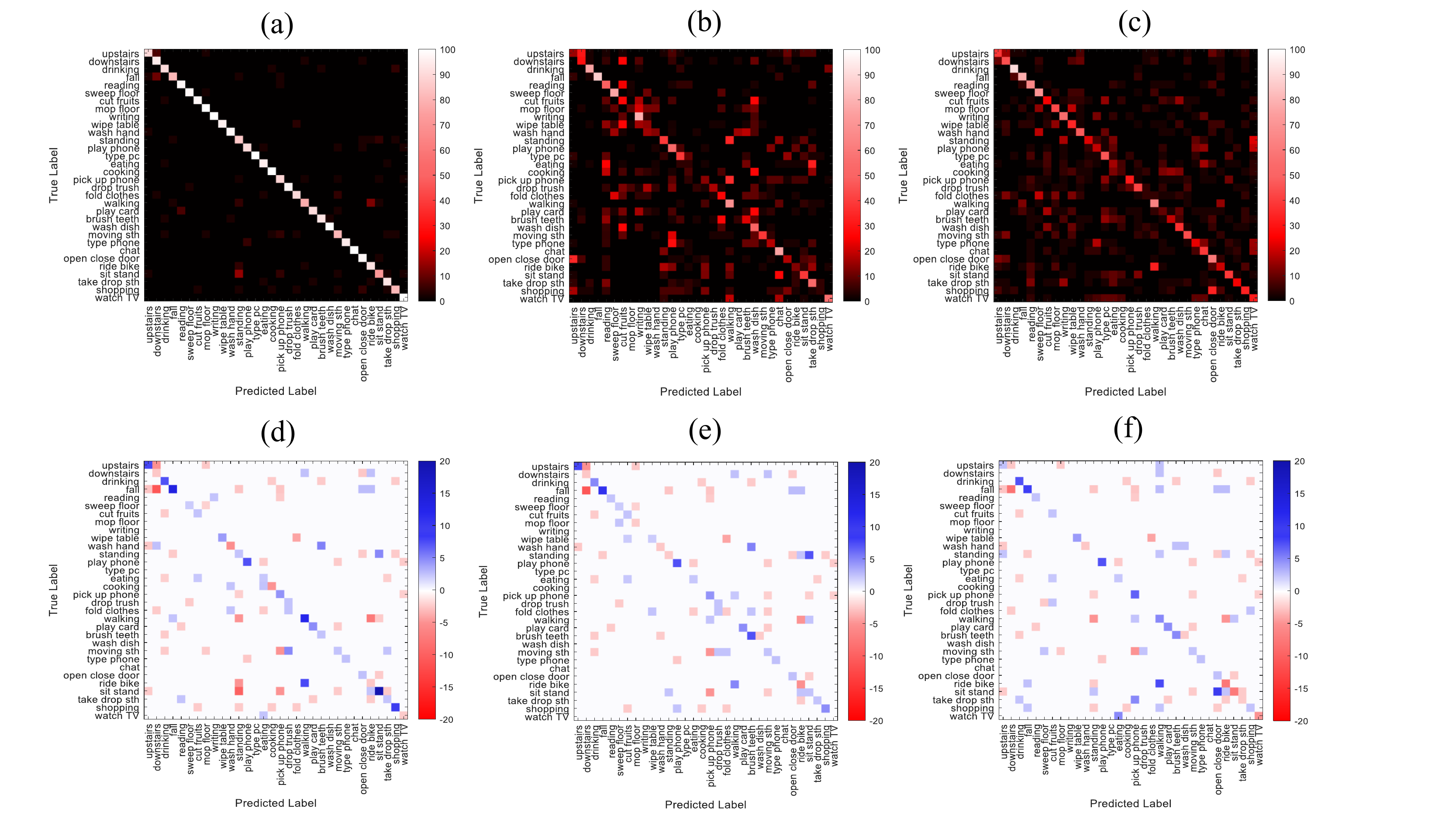}
        \vspace{-1em}
        \caption{Confusion matrices of activity recognition. The first row shows the test results for three uni-modal networks: (a) RGB; (b) Acceleration; (c) Gyroscope. The second row demonstrates the difference between the multi-modal combination networks and the RGB network: (d) Difference between `All' and `RGB'; (e) Difference between `RGB+Acc' and `RGB'; (f) Difference between `RGB+Gyro' and `RGB'.}
        \label{cfm} 
\end{figure*}

 \textbf{Multi-modal egocentric activity recognition.} 
We evaluate the multi-modal egocentric activity recognition on UESTC-MMEA-CL with different modal combinations using the base architecture in Fig.~\ref{TBN}. The results are summarized in Table~\ref{T3}. For the uni-modal recognition, RGB network achieves the most prominent performance compared with acceleration and gyroscope network. Fusing the two modalities of sensor data, the average class precision is 59.9\%, which achieves significant improvements of more than 56\% compared to the uni-modality `Acc' and `Gyro'. The improvements of the modal combination methods `RGB+Acc' and `RGB+Gyro' on RGB are not great but also obvious. When fusing all modalities together, `RGB+Acc+Gyro' achieves the highest recognition accuracy. 

In order to demonstrate the recognition performance of different modality and modal combinations on each activity class, we present the confusion matrices which are shown in Fig.~\ref{cfm}. As shown in Fig.~\ref{cfm}(d), with the help of the two types of motion-based sensor data, recognition accuracy of activities such as `upstairs', `drinking', `fall', `walking', `sit-stand' and `shopping' is obviously improved. Fig.~\ref{cfm}(e) and Fig.~\ref{cfm}(f) prove that `RGB+Acc' and `RGB+Gyro' also perform well but are not as good as `All'. 

\begin{figure*}
    \centering
    \includegraphics[width=17cm]{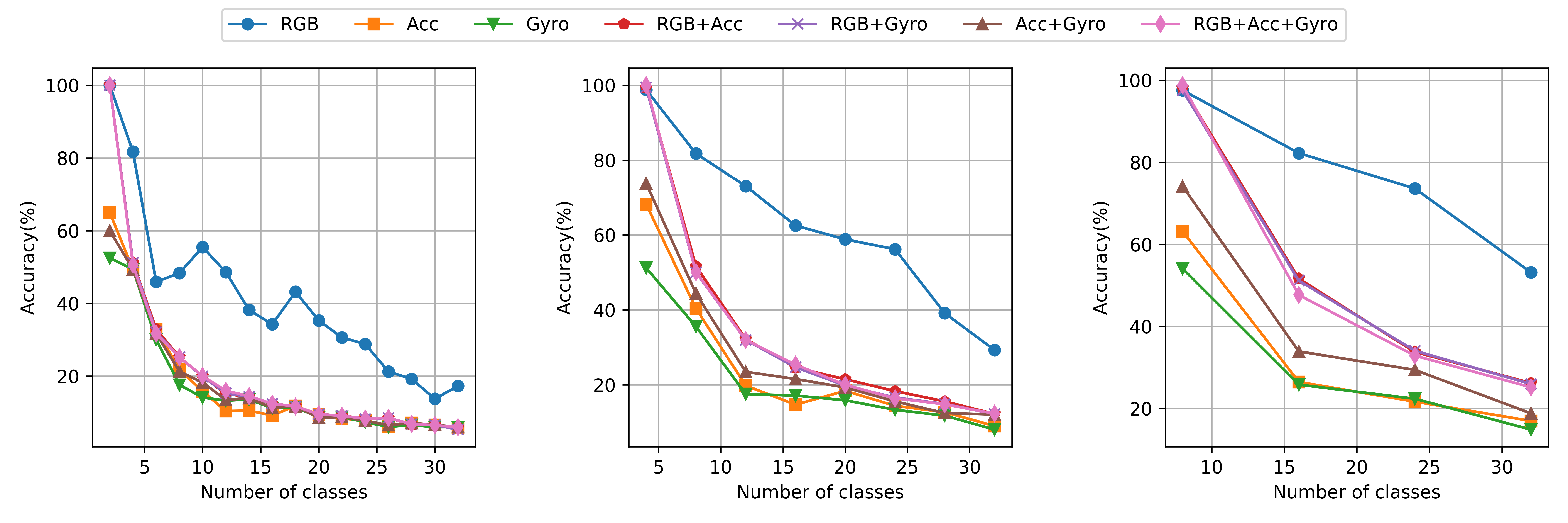}
    \caption{Fine-tune results on UESTC-MMEA-CL with $N/S = 2$ (left), 4 (middle) and 8 (right).}
    \label{result_finetune}
\end{figure*}

\textbf{Catastrophic forgetting.} Deep neural networks often suffer from catastrophic forgetting in continual learning tasks. In order to demonstrate the catastrophic forgetting in the context of continual learning for multi-modal activity recognition, the straightforward solution fine-tune is evaluated on UESTC-MMEA-CL with three incremental settings and two metrics A and F that are introduced in section \ref{metrics}. As shown in Fig.~\ref{result_finetune}, with the increase of incremental tasks, the recognition accuracy of fine-tune on different modalities and modal combinations decrease dramatically. Moreover, when using the sensor data the model suffers from more serious catastrophic forgetting than using RGB only. The average accuracy and forgetting of fine-tune with incremental setting $N/S = 4$ are demonstrated in the first column of Table~\ref{A and F}. It can be observed that uni-modal network `RGB' maintains a low forgetting rate while uni-modal network `Acc' suffers from very severe forgetting of previously learned activities. Although `Gyro' performs not well, the average forgetting of `Gyro' is not as high as `Acc' due to the low accuracy of `Gyro' at the first incremental step. It can be seen that multi-modal combinations `RGB+Acc', `RGB+Gyro', and `All' (`RGB+Acc+Gyro'), don't add gain to uni-modal network `RGB' in the continual learning. Instead, the catastrophic forgetting problem of fine-tune is aggravated with the addition of complementary sensor data.

\begin{figure*}
    \centering
    \includegraphics[width=17.8cm]{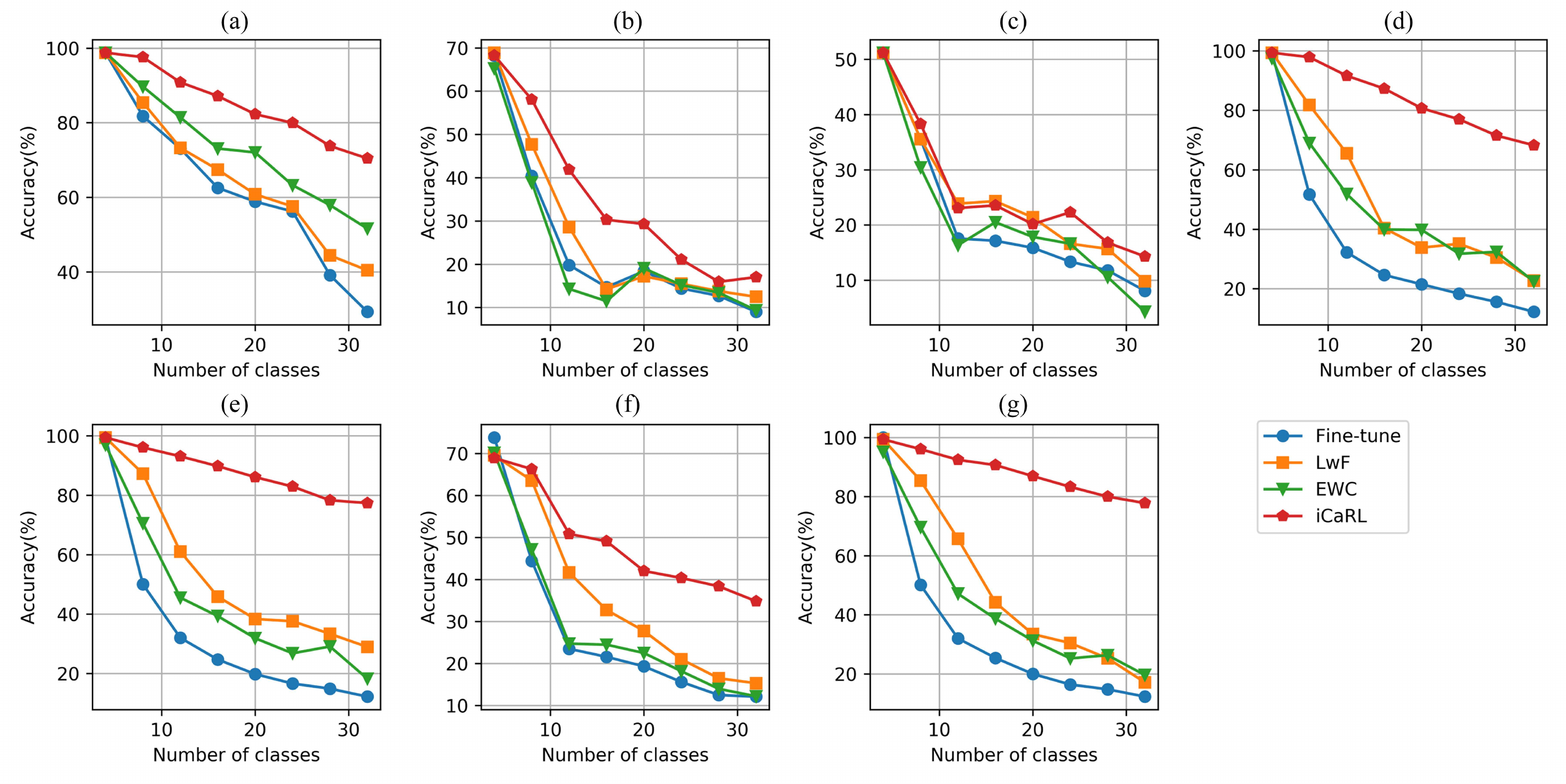}
    \caption{Multi-modal continual learning performance on UESTC-MMEA-CL with $N/S=4$. Three continual learning methods and fine-tune are evaluated on our dataset using different modalities and modal combinations: (a) RGB; (b) Acc; (c) Gyro; (d) RGB+Acc; (e) RGB+Gyro; (f) Acc+Gyro; (g) RGB+Acc+Gyro.}
    \label{result_4}
\end{figure*}

\begin{table*}[]
\caption{Average accuracy (A) and average forgetting (F) of continual learning strategies and fine-tune on UESTC-MMEA-CL ($N/S = 4$). Note that $\uparrow$ indicates the higher the better and vice versa. }
\label{A and F}
\resizebox{\textwidth}{21.6mm}{
\setlength{\tabcolsep}{4.2mm}
\begin{tabular}{l|cc|cc|cc|cc}
\toprule[1pt]
 & \multicolumn{2}{c|}{Fine-tune} & \multicolumn{2}{c|}{iCaRL \cite{Rebuffi17}} & \multicolumn{2}{c|}{EWC \cite{Kirkpatrick17}} & \multicolumn{2}{c}{LwF \cite{Li18}} \\ \cline{2-9} 
             & A$\uparrow$    & F$\downarrow$    & A$\uparrow$    & F$\downarrow$    & A$\uparrow$    & F$\downarrow$    & A$\uparrow$    & F$\downarrow$     \\ \hline
RGB          & 29.3 & 64.5 & 70.4 & 32.1 & 51.6 & 35.4  & 40.4 & 51.8  \\
Acc          & 9.0    & 86.3 & 17.0   & 67.4 & 9.4  & 49.6  & 12.5 & 20.7  \\
Gyro         & 8.1  & 68.8 & 14.3 & 58.1 & 4.3  & 42.8  & 9.8  & 33.7  \\
RGB+Acc      & 12.2 & 99.0   & 68.2 & 34.8 & 22.4 & 24.3 & 22.7 & 44.6 \\
RGB+Gyro     & 12.2 & 98.9 & 77.4 & 34.2 & 18.2 & 18.6  & 29.0   & 49.0    \\
Acc+Gyro     & 12.1 & 83.3 & 34.8 & 56.9 & 12.2 & 66.1  & 15.3 & 15.7  \\
RGB+Acc+Gyro & 12.3 & 99.0   & \textbf{77.8 }& 33.5 & 19.6 & 29.9  & 17.1 & 49.4  \\ 
\bottomrule[1pt]
\end{tabular}
}
\end{table*}

\textbf{Evaluation with continual learning strategies.} To overcome catastrophic forgetting, we transfer the popular continual learning methods iCaRL \cite{Rebuffi17}, EWC \cite{Kirkpatrick17} and LwF \cite{Li18} to the continual multi-modal activity recognition. Fig.~\ref{result_4}(a) demonstrates the suppression of catastrophic forgetting by continual learning strategies with RGB uni-modal network. With the help of exemplar replay, iCaRL effectively alleviates the forgetting problem while the effectiveness of exemplar-free strategies EWC and LwF is not so noticeable. As shown in Fig.~\ref{result_4}(b) and (c), these continual learning strategies do not produce the same effect on alleviating forgetting of the acceleration and gyroscope uni-modal networks as the RGB network. As listed in the second and third rows of Table~\ref{A and F}, the average accuracy of the sensor networks is below 20\% even if continual learning strategies are adopted.

Fig.~\ref{result_4}(d)-(g) present the recognition accuracy of the multi-modal combination networks. When combining the sensor data, the replay-based iCaRL can implicitly exploit the multi-modal complementary information to reduce the forgetting rate of RGB network. As shown in Table~\ref{A and F}, `iCaRL-RGB+Acc+Gyro' achieves the highest average accuracy 77.8\% and a relatively low forgetting rate 33.5\%. `iCaRL-RGB+Gyro' also perform well with average accuracy 77.4\% compared with 70.4\% of `iCaRL-RGB'. Compared with fine-tune, exemplar-free strategies EWC and LwF can also suppress the model forgetting, but not obvious as iCaRL.

\subsection{Discussion}
\textbf{Fusion of multi-modal data.} In our work, we use TBW-like midfusion to aggregate features from different modalities, while it can not be ignored that an early fusion or late fusion way will receive different performances against catastrophic forgetting. Exploring a more reasonable manner to fuse and align the multi-modal data deserves further study.

\textbf{Catastrophic forgetting of sensor modalities.} As shown in Fig.~\ref{result_4}(b), (c), and (f), continual learning using sensor modalities performs poorly if RGB is unavailable and the forgetting problem is severer than using RGB. This phenomenon may be closely related to the network architecture. 

\textbf{Continual leaning without exemplar}. In this paper, three popular continual learning strategies, i.e., exemplar-based method iCaRL and exemplar-free methods LwF and EWC, are evaluated on UESTC-MMEA-CL. The experimental results indicate that exemplars can effectively alleviate the forgetting problem in multi-modal networks. However, in practical applications, especially in services involving privacy, it is not always available to select and store exemplars. Therefore, studying how to improve the catastrophic forgetting problem of multi-modal networks (especially sensor networks) under exemplar-free case will be an vital research direction in the future. 

\section{Conclusion}
 In this paper, we propose a multi-modal egocentric dataset, named UESTC-MMEA-CL, for continual activity recognition task. UESTC-MMEA-CL contains video, acceleration, and gyroscope data of 32 daily activity classes. Compared to the existing multi-modal datasets, UESTC-MMEA-CL provides not only vision data with auxiliary inertial sensor data but also abundant categories for the purpose of continual learning research. Besides, a baseline model is presented for continual multi-modal egocentric activity recognition. We have conducted comprehensive experiments on UESTC-MMEA-CL to explore catastrophic forgetting of multi-modal networks and evaluate four baseline methods to address this problem. Finally, we have given some potential research directions for future research. We hope our multi-modal egocentric dataset can facilitate future studies on multi-modal first-person activity recognition as well as continual learning in wearable applications.

\bibliographystyle{IEEEtran}
\bibliography{IEEEabrv,Multimodal_Egocentric_Dataset}
\begin{comment}

\end{comment}

%Biography text here without a photo.

%\begin{IEEEbiography}{IEEE Publications Technology Team}
%In this paragraph you can place your educational, professional background and research and other interests.\end{IEEEbiography}

\end{document}